\documentclass[conference]{IEEEtran}
\usepackage{times}

\usepackage[numbers]{natbib}
\usepackage{multicol}
\usepackage[bookmarks=true]{hyperref}
\usepackage{amsmath}
\usepackage{amssymb}
\usepackage{bm}
\usepackage{xcolor}
\usepackage{booktabs}
\usepackage{multirow}
\usepackage{makecell}
\usepackage[table]{xcolor}
\usepackage{siunitx}
\newcommand{\ci}[1]{\tiny{\textcolor{gray}{$\pm #1$}}}

\usepackage{graphicx}      
\usepackage{wrapfig}       
\usepackage{float}         
\usepackage{stfloats}      
\usepackage{capt-of}       
\usepackage[font=footnotesize,labelfont=bf]{caption}      

\begin{document}

\title{\textbf{Pro-HOI}: \textbf{P}erceptive \textbf{Ro}ot-guided \textbf{H}umanoid-\textbf{O}bject \textbf{I}nteraction}


    
    


%
\author{\authorblockN{Yuhang Lin\textsuperscript{\rm 1,2} \quad
Jiyuan Shi\textsuperscript{\rm 1} \quad
Dewei Wang\textsuperscript{\rm 1,3} \quad
Jipeng Kong\textsuperscript{\rm 1,4} \quad \\
Yong Liu\textsuperscript{\rm 2} \quad
Chenjia Bai\textsuperscript{\rm 1}\authorrefmark{2} \quad
Xuelong Li\textsuperscript{\rm 1}\authorrefmark{2}
}
\authorblockA{
\textsuperscript{\rm 1}Institute of Artificial Intelligence (TeleAI),  China Telecom \quad 
\textsuperscript{\rm 2}Zhejiang University \\ 
\textsuperscript{\rm 3}University of Science and Technology of China \quad 
\textsuperscript{\rm 4}ShanghaiTech University \\
\textsuperscript{\dag{}}Corresponding author
}
}

\twocolumn[{%
\renewcommand\twocolumn[1][]{#1}%
\maketitle
\thispagestyle{empty}
\pagestyle{empty}
\begin{center}
    \vspace{-0.2cm}
    \centering
    \captionsetup{type=figure}
     \includegraphics[width=0.9\textwidth]{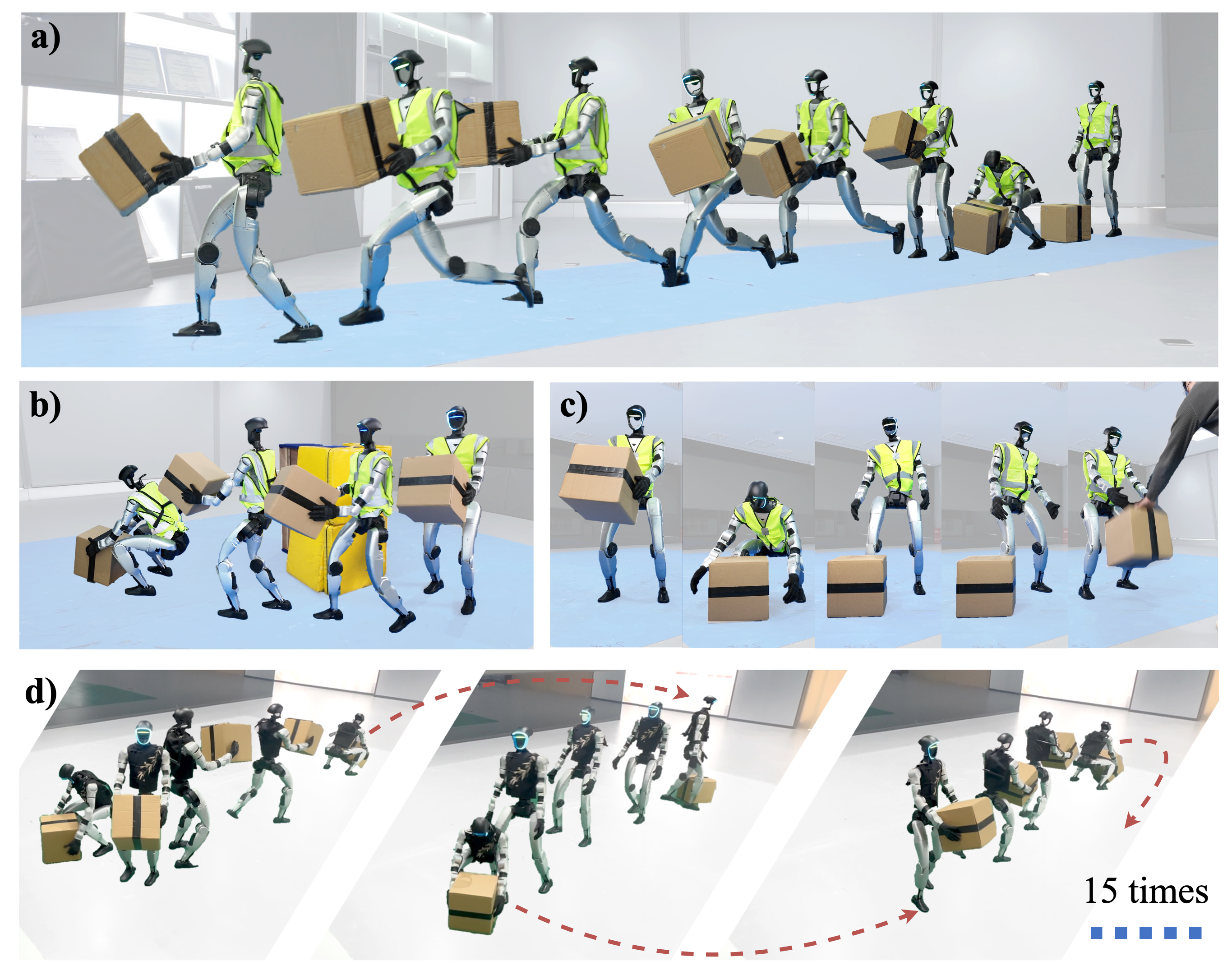}
    \caption{\textbf{Pro-HOI} enables humanoid robots to perform generalizable and robust whole-body object interaction using exclusively onboard sensing and computing resources on a Unitree G1. a) \textbf{Stylized Behaviors}: Guided by motion tracking rewards, the robot exhibits stylized skills, such as running while carrying. b) \textbf{Navigation \& Obstacle Avoidance}: The root-guided interface facilitates seamless integration with high-level planners for simultaneous carrying and obstacle avoidance. c) \textbf{Autonomous Re-grasping}: The robot autonomously detects an accidental drop, utilizing a digital twin module to enable state estimation and retrieval even when the object falls out of the camera's field of view. d) \textbf{Continuous Operation}: The robot successfully completes over 15 consecutive carrying cycles, demonstrating long-horizon robustness.}
    \label{fig:teaser}
\end{center}
}]

\begin{abstract}
Executing reliable Humanoid-Object Interaction (HOI) tasks for humanoid robots is hindered by the lack of generalized control interfaces and robust closed-loop perception mechanisms. In this work, we introduce \underline{P}erceptive \underline{Ro}ot-guided \underline{H}umanoid-\underline{O}bject \underline{I}nteraction, Pro-HOI, a generalizable framework for robust humanoid loco-manipulation. First, we collect box-carrying motions that are suitable for real-world deployment and optimize penetration artifacts through a Signed Distance Field loss. Second, we propose a novel training framework that conditions the policy on a desired root-trajectory while utilizing reference motion exclusively as a reward. This design not only eliminates the need for intricate reward tuning but also establishes root trajectory as a universal interface for high-level planners, enabling simultaneous navigation and loco-manipulation. Furthermore, to ensure operational reliability, we incorporate a persistent object estimation module. By fusing real-time detection with Digital Twin, this module allows the robot to autonomously detect slippage and trigger re-grasping maneuvers.  Empirical validation on a Unitree G1 robot demonstrates that Pro-HOI significantly outperforms baselines in generalization and robustness, achieving reliable long-horizon execution in complex real-world scenarios. The project page is available on \href{https://pro-hoi.github.io/}{project page}.
\end{abstract}

\IEEEpeerreviewmaketitle
\begin{table*}[tp]
    \centering
    \caption{Capability comparison across different methods} 
    \label{tab:capability_comparison}
    \centering 
    \footnotesize
    \setlength{\tabcolsep}{8pt}
        \begin{tabular}{lccccccccc} 
        \toprule
        Method        
        & \makecell{No External \\Devices} 
        & \makecell{Whole-Body \\Control}
        & \makecell{Generalizable \\Interaction}
        & \makecell{Controlable}
        & \makecell{Failure \\Recovery} \\ 
        \midrule
        OmniRetarget\cite{omniretarget} & $\checkmark$ & $\checkmark$ & $\times$ & $\times$ & $\times$ \\
        HDMI\cite{hdmi} & $\times$ & $\checkmark$ & $\times$ & $\times$ & $\times$ \\
        DemoHLM & $\checkmark$ & $\checkmark$ & $\checkmark$ & $\times$ & $\times$ \\
        PhysHSI\cite{physhsi} & $\checkmark$ & $\checkmark$ & $\checkmark$ & $\times$ & $\times$ \\
        Falcon\cite{zhang2025falcon} & $\checkmark$ & $\checkmark$ & $\checkmark$ & $\checkmark$ & $\times$ \\
        OURS & $\checkmark$ & $\checkmark$ & $\checkmark$ & $\checkmark$ & $\checkmark$ \\
        \bottomrule
        \end{tabular}
    \vspace{-0.4cm}
\end{table*}


\section{Introduction}
Humanoid robots, characterized by their high degrees of freedom (DoFs) and anthropomorphic configurations, are widely viewed as the ideal platform for general-purpose robots. While state-of-the-art methods have demonstrated remarkable capabilities for performing highly dynamic motions \cite{beyondmimic, kungfubot,kungfubot2,any2track} and traversing complex terrain \cite{humanoidparkour,PIM,more}, executing whole-body manipulation tasks, such as box carrying, remains a significant challenge. This difficulty arises from the rigorous demands of maintaining balance under external loads, closed-loop perception for robust interaction with objects \cite{gu2025humanoid}, and the extensibility to integrate with high-level planners \cite{ding2024quar}. Bridging the gap from isolated locomotion primitives to integrated perception-control frameworks is essential for achieving reliable carrying capabilities and represents a critical milestone in transitioning humanoid robots from laboratory demonstrations to practical applications in human-centric environments.

Existing works on humanoid robot carrying tasks face various limitations. Traditional model-based methods often require task-specific design of optimization objectives and are limited to precise dynamics modeling \cite{figueroa2020dynamical, adu2023exploring, liu2025opt2skill}, making it difficult to generalize to unknown scenarios. Reinforcement Learning (RL) has emerged as a scalable alternative for discovering complex contact-rich behaviors \cite{hassan2023synthesizing,xu2025intermimic,hou2025efficient}. However, end-to-end RL approaches often necessitate laborious reward engineering, requiring distinct parameter tuning for different task phases to ensure convergence \cite{wococo}. Recent advancements have attempted to address naturalness and generalization by incorporating Adversarial Motion Priors (AMP) \cite{amp} and a coarse-to-fine perception scheme \cite{physhsi}. HDMI \cite{hdmi} proposes to learn interaction skills from videos and presents a unified interaction task design, yielding a unified HOI training framework. However, this method not only relies on optical motion capture systems that are difficult to migrate, but also limits the location of object interactions to the training motion. What's more, existing approaches often lack a generalized command interface for high-level planners \cite{xie2023drl} and fail to incorporate closed-loop perception during task execution \cite{omniretarget}. For instance, if a robot drops a box while carrying, a practical system should enable the robot to autonomously detect the failure and pick it up. These limitations render current methods impractical for real-world carrying tasks.

In this work, we present \textbf{Pro-HOI} (\textbf{P}erceptive \textbf{Ro}ot-guided \textbf{H}umanoid-\textbf{O}bject \textbf{I}nteraction), a generalizable and robust RL controller for HOI task. First, we decouple the motion tracking objective from the observation space, conditioning the policy solely on the desired root trajectory while utilizing whole-body tracking exclusively as a reward. This enables the policy to learn the dynamic correlation between root trajectory and task states, allowing for generalizable object interaction without requiring full-body reference motions. Furthermore, this design establishes the root trajectory as a universal interface, facilitating seamless integration with diverse high-level planners for executing carrying, navigation, and obstacle avoidance tasks simultaneously. Second, to address real-world uncertainties such as accidental drops, we introduce a persistent object estimation module. Combining real-time object detection with Digital Twin, this module enables the robot to detect drops and re-acquire objects autonomously, thereby substantially enhancing system reliability. Table~\ref{tab:capability_comparison} presents a qualitative comparison between our framework and state-of-the-art HOI methods. While early approaches like OmniRetarget~\cite{omniretarget} and HDMI~\cite{hdmi} established the foundation for whole-body control, they are limited by either poor generalization or reliance on external devices. Recent imitation-based works, such as PhysHSI~\cite{physhsi} and Falcon~\cite{zhang2025falcon}, have significantly improved interaction generalizability and controllability. However, a critical gap remains: none of these baselines account for execution robustness. In contrast, our method stands out as the only holistic solution that integrates controllable, generalizable interaction with a distinct failure recovery mechanism, ensuring reliable deployment without the need for external instrumentation.

To evaluate our approach, we conducted extensive experiments on the box carrying task with various practical challenges, such as continuous multi-cycle transport of boxes with diverse locations, carrying with obstacle avoidance, and autonomous re-grasping under disturbances. The results confirm that Pro-HOI not only supports seamless high-level integration but also achieves superior robustness, enabling reliable long-horizon task execution.

In summary, the main contributions of this work are as follows:
\begin{itemize}
    \item We propose a root trajectory guided RL framework for HOI task. This framework achieves generalizable humanoid HOI capabilities without requiring extensive reference motions or intricate reward engineering. Furthermore, the root trajectory serves as a flexible interface that seamlessly integrates with high-level planners to enable simultaneous navigation and obstacle avoidance. 
    \item We introduce a persistent object estimation module that combines real-time object detection with Digital Twin. This module allows autonomous re-grasping of dropped objects, which significantly enhances system robustness and practicality in real-world scenarios.
    \item We deploy the complete system on a Unitree G1 robot, with all modules running on a single onboard Jetson NX. Extensive real-world experiments validate the generalization, scalability, and robustness of our approach.
\end{itemize}

\section{Related Works}
\subsection{Reinforcement Learning for Humanoid Control}
With large-scale simulation training and sim-to-real transfer techniques, RL empowers humanoid robots to achieve robust locomotion \cite{radosavovic2024real, xie2025humanoid, PIM}, highly agile maneuvers \cite{humanoidparkour, jump}, and recovery behaviors \cite{he2025learning, huang2025learning}. 
To enable robots to perform human-like behaviors, mimic-based approaches that imitate human demonstrations \cite{peng2018deepmimic} have been extended to highly anthropomorphic and dynamic humanoid whole-body control \cite{he2025asap, kungfubot, beyondmimic, any2track, kungfubot2, omniretarget}. Despite demonstrating remarkable capabilities, these methods exhibit limited generalization, particularly for real-world HOI tasks that demand extensive adaptability to diverse scenarios. Alternative paradigms like Adversarial Motion Priors (AMP) \cite{more, physhsi} mitigate the need for precise tracking by learning style distributions from datasets.
However, these methods typically suffer from a trade-off between style fidelity and task generalization.

Distinguished from prior works, we propose a root-guided HOI training paradigm. By decoupling the root commands from the motion tracking rewards, the policy only exhibits anthropomorphic behaviors but also possesses high generalizability.

\subsection{Humanoid-Object Interaction}
Humanoid-object interaction (HOI) \cite{pan2025tokenhsi, wang2025sims} presents a compounded challenge due to the high-dimensional state space and complex contact dynamics involved, particularly in real-world deployments. Early approaches utilizing Trajectory Optimization (TO) \cite{ruscelli2020multi, figueroa2020dynamical, liu2025opt2skill} can generate physically feasible motions for specific tasks but often suffer from limited generalization to unseen scenarios. 

Recent RL-based approaches for HOI have demonstrated enhanced robustness and generalization capabilities \cite{wococo, hdmi, physhsi, zhang2025falcon, yin2025visualmimic}. Through meticulous reward and task design, RL-based controllers can acquire whole-body manipulation skills involving specific contact sequences \cite{wococo}. However, such methods necessitate intricate reward engineering and, due to the absence of reference trajectories, often struggle to constrain the stylistic quality of the learned motions. A promising alternative involves learning interaction skills directly from large-scale human videos \cite{hdmi}. While data acquisition in this paradigm is scalable, the approach relies on external Motion Capture (MoCap) systems for deployment and often lacks generalization regarding object positioning. VisualMimic\cite{yin2025visualmimic} employs a hierarchical whole-body control framework to train a humanoid robot for specific HOI tasks; however, the learned policy is limited to interactions with objects that remain within the visual field. FALCON\cite{zhang2025falcon} achieves force-adaptive HOI through a dual-agent learning framework and can be combined with high-level planners to perform complex tasks. However, these methods are limited to scenarios where the object remains within the robot's field of view. Distinguished from the motion-tracking or hierarchical methods, PhysHSI \cite{physhsi} incorporates AMP to learn an end-to-end HOI policy and significantly enhance generalization, yet this inevitably introduces the burden of laborious task reward tuning.

In contrast, our approach decouples the target root trajectory observation from the motion tracking reward. This design enables the robot to execute generalizable HOI tasks while preserving human-like motion styles. Furthermore, the root trajectory serves as a versatile interface compatible with high-level planners, facilitating complex tasks such as navigation and obstacle avoidance. Additionally, by leveraging a digital twin module to enable state estimation for out-of-view objects during deployment, our system successfully completed over 15 continuous box-carrying cycles in the real world, demonstrating exceptional generalization and robustness.

\begin{figure*}
    \centering
    \includegraphics[width=\linewidth]{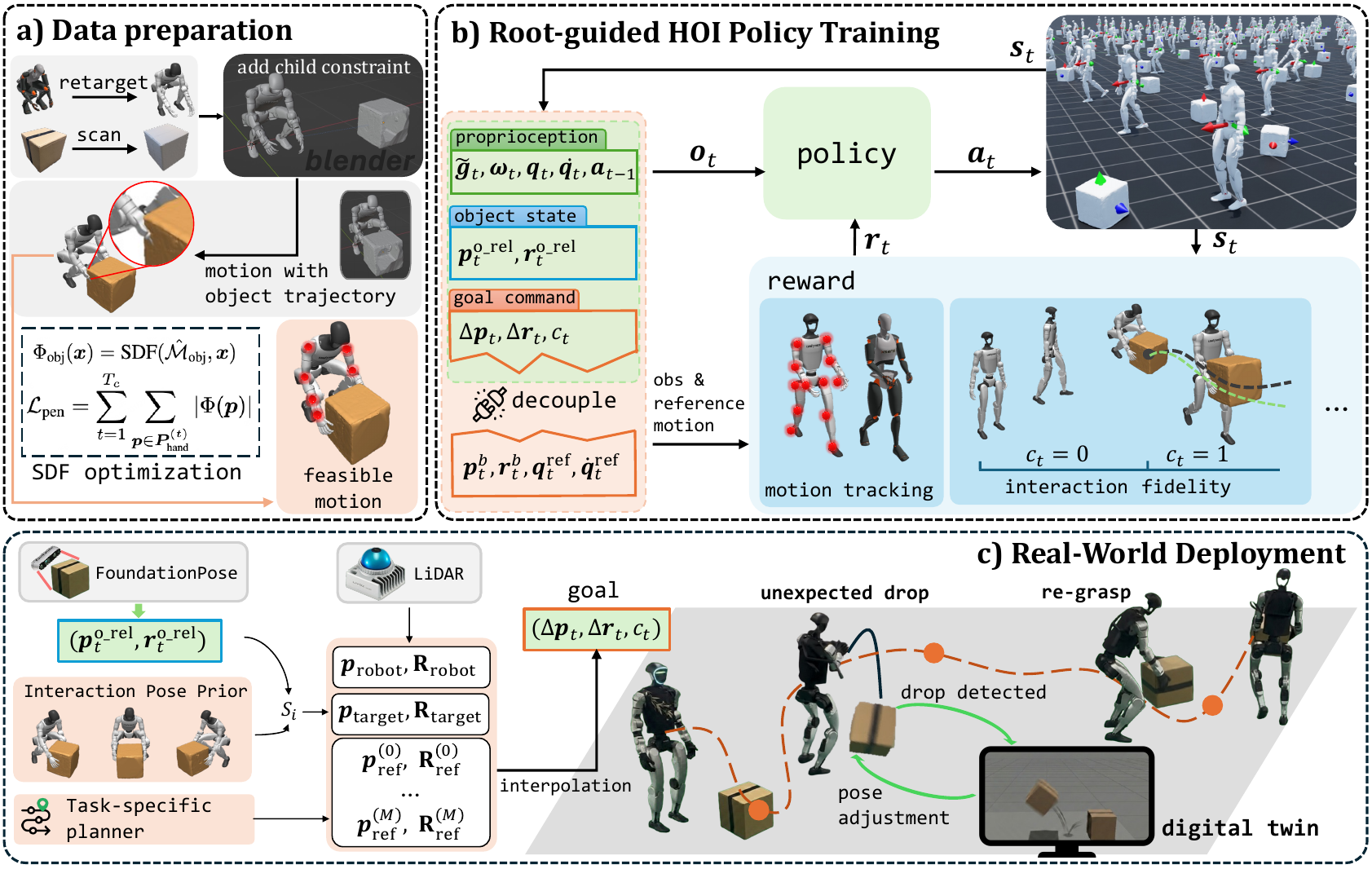}
    \caption{\label{fig:}\textbf{Overview of  Pro-HOI.} a) \textbf{Data Preparation}: We frist collect human motion clips using the mocap system,  and augment them with object geometries in \textit{Blender}. Then we use SDF-based optimization to generate physically feasible reference motions. b) \textbf{Root-Guided Policy Learning}: The RL policy is trained to perform whole-body interaction skills conditioned on the desired root trajectory, while utilizing the reference motion as rewards. c)\textbf{ Real-world Deployment}: We integrate FoundationPose for 6D object pose estimation and FAST-LIO2 \cite{xu2022fast} for root pose estimation, combined with the Interaction Pose Prior and a task specific planner to generate target root trajectories. The full stack executes entirely onboard a Jetson NX with a D435i camera and a Mid-360 LiDAR, achieving robust sim-to-real transfer.}
\vspace{-0.4cm}
\end{figure*}

\section{Method}
\subsection{Problem Definition}
We formulate humanoid-object interaction as a goal-conditioned reinforcement learning problem, where the agent interacts with objects in a physical environment according to a policy $\pi$ to maximize cumulative reward. We adopt the Unitree G1 robot as the agent, which has 29 degrees of freedom (DoF) in total. At each timestep $t$, the policy receives the state $\bm s_t$ and target state $\bm g_t$, and outputs an action $\bm{a} \in \mathbb{R}^{29}$, which represents the target joint positions and subsequently converted into motor torques via a PD controller. State transitions are governed by the environment dynamics $p(\bm s_{t+1}|\bm s_t, \bm a_t)$, and A dense reward function $r(\bm s_t,\bm g_t,\bm a_t)$ is employed for task performance evaluation and regularization. The optimization objective is to maximize the expected discounted return: $J = \mathbb{E} \left[ \sum_{t=0}^{T-1} \gamma^t r_t \right]$. 

The policy is trained using Proximal Policy Optimization (PPO) \cite{schulman2017proximal} in an asymmetric Actor-Critic architecture \cite{pinto2017asymmetric} to ensure it is deployable on physical hardware. During training, the critic has access to privileged information (e.g., base velocity), while the actor operates on noisy observations available to the real robot.

\subsection{Data Preparation and Optimization}
While numerous HOI datasets exist \cite{li2023object,bhatnagar22behave}, their high-dynamic motions (e.g., sharp bending) are often difficult for a humanoid robot to imitate, thereby increasing the risk of overheating. To improve interaction performance, we first collected a custom dataset of box-carrying human motions in the SMPL \cite{smplx} format using an Xsens inertial motion capture system. We then retargeted the SMPL motions to the humanoid robot using General Motion Retargeting (GMR) \cite{joao2025gmr}. Without capturing object state, existing HOI retargeting methods \cite{omniretarget} are inapplicable. To address this, we manually associated the object with the robot's end-effector using the `Child Of' constraint in Blender \cite{Blender2024}.

Precise humanoid–object alignment is crucial for mimic-based learning frameworks. We formulated a penetration loss based on SDF \cite{park2019deepsdf,ye2025dex1b} to resolve mesh penetration artifacts between the robot's end-effector and the object. 

Specifically, we utilize voxel-based downsampling to obtain simplified meshes $\hat{\mathcal{M}}_\text{hand}$ and $\hat{\mathcal{M}}_\text{obj}$, and construct a SDF $\Phi_{\text{obj}}(\cdot)$ from the object mesh. Then, we optimize the robot's upper-body joints position $\bm{q}_{\text{upper}}$ over the contact phase $T_c = \left[\phi_1,\phi_2\right]$, which is computed by the linear velocity of objects. The penetration loss is defined as:
\begin{equation}
    \mathcal{L}_\text{pen} = \sum_{t=\phi_1}^{\phi_2} \sum_{\bm{p} \in \bm{P}^\text{hand}_{t}} \left\vert \Phi_{\text{obj}}(\bm{p}) \right\vert
\end{equation}
where $\bm{P}^\text{hand}_{t}$ represents the hand vertices at timestep $t$ in the object's coordinate frame. We minimize $\mathcal{L}_\text{pen}$ using the Adam optimizer~\cite{kingma2015adam} to keep the hands in contact with the object's surface.

\subsection{Root-guided HOI Policy Training}
\subsubsection{HOI Process Encoding} 
A core challenge in mimic-based HOI is the trade-off between motion fidelity and generalization. Existing imitation approaches often include full-body reference motions (e.g., joint positions and velocity) \cite{beyondmimic} or scalar phase variables (e.g., $\phi \in [0,1]$) \cite{he2025asap} in the policy observation. However, directly observing high-dimensional reference motions imposes a heavy data burden for learning the intricate mapping between body poses and object locations \cite{luo2025sonic}. Since interaction mechanics vary drastically with object positioning, achieving exhaustive dataset coverage is practically infeasible. Conversely, low-dimensional phase variables lack sufficient spatial context to guide complex, multi-stage tasks like box carrying.

To bridge this gap, we propose a root-guided observation paradigm. By leveraging the desired root movement trend, we effectively encode the task's semantic stages. In the case of carrying boxes, vertical movement means squatting or standing, while horizontal movement means locomotion or transport. Coupling this spatial trend with a binary desired contact state $c_t \in \{0,1\}$ forms a minimalist yet sufficient observation space. Desired contact state $c_t$ represents whether the robot's end-effectors should be in contact with the object from reference motions. This design compels the policy to learn the correlations between desired movement, contact states, and distinct task phases via reward signals, rather than merely mimicking reference motions. Notably, this streamlined design exhibits high data efficiency, requiring only a limited set of HOI demonstrations for effective training. Furthermore, utilizing the root trajectory as an observation facilitates seamless integration with high-level path planners, thereby enabling the execution of complex object interaction tasks.

Specifically, we define the above observation state as the goal $\bm{g}_t=\left[ \Delta \bm p_t^\text{root}, \Delta \bm r_t^\text{root}, c_t  \right]$. The $\Delta \bm p_t^\text{root}$ and $\Delta \bm r_t^\text{root}$ represent the deviation in root position and orientation between the current state and the expected state. 

\subsubsection{State Space} 
The critic and the actor operate on different information sets. At each timestep $t$, the critic has access to the full state $\bm{s}_t$, which comprises the proprioceptive state, the object state $\bm{s}^\text{o}_t$, and the goal $\bm{g}_t$:
\begin{equation}
    \bm{s}_t = \left[
    \bm p_t^\text{body}, \bm r_t^\text{body}, \tilde{\mathbf{g}}, \bm v_t, \bm \omega_t, \bm q_t, \dot{\bm q}_t, \bm a_{t-1}, \bm p^\text{o\_rel}_t, \bm r^\text{o\_rel}_t,\bm{g}_t
    \right]
\end{equation}
where $\bm p_t^\text{body} \in \mathbb{R}^{14 \times 3}$ and $\bm r_t^\text{body} \in \mathbb{R}^{14 \times 6}$ denote the positions and rotations of the robot's key bodies. The root state consists of projected gravity $\tilde{\mathbf{g}} \in \mathbb{R}^3$, linear velocity $\bm v_t \in \mathbb{R}^3$, and angular velocity $\bm \omega_t \in \mathbb{R}^3$. The joint state includes positions $\bm q_t \in \mathbb{R}^{29}$, velocities $\dot{\bm q}_t \in \mathbb{R}^{29}$, and the previous action $\bm a_{t-1} \in \mathbb{R}^{29}$. The state of an object consists of its position $\bm p^\text{o\_rel}_t \in \mathbb{R}^3$ and rotation $\bm r^\text{o\_rel}_t\in\mathbb{R}^6$ in the robot coordinate system.

In contrast, the actor $\pi$ receives a partial observation $\bm{o}_t$ to ensure deployment feasibility. Specifically, $\bm{o}_t$ excludes the privileged information difficult to estimate on physical hardware—namely, the robot's key body poses and the root linear velocity $\bm v_t$. Thus, the observation is defined as a subset: $\bm{o}_t = \bm{s}_t \setminus \{ \bm p_t^\text{body}, \bm r_t^\text{body}, \bm v_t \}$. Furthermore, to bridge the sim-to-real gap, we inject Gaussian noise into the policy's observations to simulate sensor uncertainty, whereas the critic operates on the noise-free state to ensure stable value estimation. 

\subsubsection{Reward Terms} 
Without task-specific reward design, we utilize simple tracking-based reward terms $\mathcal{R}(\cdot,\cdot)$ to guide the training process. During training, for each episode, we randomly sample a reference HOI motion state $\bm {s}_t^\text{ref}$ from the dataset. The robot and the object are initialized to the starting state of the sequence. At each timestep $t$, we compute the total reward $r_t$ as the sum of two components: 
\begin{equation}
    r_t = \mathcal{R}(\bm {s}_t,\bm {s}_t^\text{ref}) + \mathcal{P}_{\text{reg}}(\bm s_t,\bm a_t)
\end{equation}
Specifically, the tracking reward $\mathcal{R}(\bm s_t,\bm {s}_t^\text{ref})$ minimizes deviations in the global root pose and incentivizes the agent to replicate the reference joint kinematics, thereby implicitly guiding the necessary whole-body coordination. Regarding the object, the reward prioritizes manipulation fidelity; it encourages precise tracking of the reference trajectory and enforces consistency between the simulated and reference contact states.
Finally, the penalty term $\mathcal{P}_{\text{reg}}(\bm s_t^\text{p},\bm a_t)$ serves as a regularization component. It penalizes large temporal variations in actions to ensure control smoothness, discourages joint limit violations, and prevents undesired collisions.

\subsubsection{Sim-to-Real Transfer}
To ensure successful sim-to-real transfer, we employ Domain Randomization (DR) \cite{peng2018sim} during training to enhance policy robustness. Specifically, we randomize a set of physical parameters, including friction, restitution coefficients, joint position offsets, the agent's center of mass (CoM), and the object's mass. Furthermore, to simulate external disturbances, we apply random perturbations to the robot's root linear and angular velocities.

\subsection{Real-world Deployment System} 
A primary discrepancy between simulation and the real world lies in the inaccessibility of a precise global pose for both the robot and the object. Accurate acquisition of these states is critical for robust object interaction and precise transport. On one hand, relying on onboard sensors for perception presents significant challenges due to limited FoV, perception latency, and constrained computational resources. The difficulty is further exacerbated during occlusions or unconstrained motion (e.g., free-falling). On the other hand, training mimic-based policies typically relies on reference root trajectories from datasets. However, these trajectories often imply the location of objects, limiting the generalization ability of policy to interact with objects at random locations in real-world deployments. To address these challenges, we propose an object state estimation module (\ref{subsubsec:object_estimation}) to ensure acquisition of stable object states, and root trajectory planner (\ref{subsubsec:root_planner}) conditioned on object initial position and waypoints of the robot to achieve generalizable interactions.

\subsubsection{Object State Estimation Module}
\label{subsubsec:object_estimation}
To ensure robust 6D pose tracking under partial occlusion during interaction, we leverage FoundationPose \cite{wen2024foundationpose}. This unified foundation model enables stable tracking of novel objects given their mesh models without fine-tuning.

Once an object falls, it often exits the limited FoV of the onboard sensors. Treating the object as a point mass under gravity is insufficient, as it neglects geometric collisions with the robot body and the ground. To address this, we propose a Digital Twin module powered by the MuJoCo simulator \cite{todorov2012mujoco}. Leveraging continuous state estimation, we formulate a heuristic to detect manipulation failures. A drop event is identified based on the inconsistency between the contact goal and the relative object pose. Specifically, the system triggers a drop alert if the contact state remains active ($c_t=1$) while the deviation in the relative object position exceeds a safety threshold $\delta_{\text{pos}}$ over a sliding time window $N$.

This Digital Twin module runs on an external computational unit, synchronizing with the robot and object global states in real-time. Upon detecting a drop, the simulator initializes a parallel environment using the object's last known pose and velocity. It then performs high-fidelity physics stepping to simulate collision dynamics and predict the final resting pose. Based on this prediction, the robot automatically adjusts its gaze to re-acquire the object, facilitating seamless recovery and re-grasping.

\subsubsection{Root Trajectory Planner}
\label{subsubsec:root_planner}
The core observation mechanism of our policy relies on the deviation of the root pose between the current and reference states, which guides the entire interaction process. During training, the policy relies on a limited set of fixed reference trajectories. However, the reference trajectories can be flexibly defined to meet task requirements during inference, utilizing tools such as off-the-shelf 2D planners or generative motion models. 
To ensure stable object interaction, we propose a hybrid root trajectory planning method that leverages grasping priors from the dataset to guide the policy. Specifically, the planner consists of three parts: interaction pose prior, task-specific planner, and 6Dof trajectory interpolation $\xi(t)$, constrained by expected linear and angular velocity.

\paragraph{Interaction Pose Prior}
To facilitate robust object interaction, we leverage grasping priors extracted from the training dataset. Specifically, we record the relative robot-object poses at the onset of each contact phase. During deployment, upon detecting a new object, the system retrieves the optimal interaction configuration by evaluating the similarity between the currently detected relative object pose and the stored priors.

\paragraph{Task-specific planner}
Upon establishing the global interaction target, the policy can be flexibly guided by diverse trajectory planners to accommodate specific task requirements. In this work, we integrate the Timed Elastic Band (TEB) \cite{rosmann2017teb} local planner as an exemplar navigation module for executing long-horizon transport tasks with collision avoidance. Given a goal location and an obstacle map, the planner generates a smooth planar trajectory comprising a sequence of 2D positions and yaw angles. We then transform these planar waypoints into a 6-DoF trajectory to serve as the desired root poses.

\paragraph{6Dof trajectory interpolation}
To generate a continuous reference root trajectory for the policy, we apply 6-DoF interpolation to the constructed waypoint sequence. This process synthesizes the robot's current state, specific interaction targets, and the planner-generated trajectory into a unified time-parameterized trajectory, while simultaneously outputting the expected contact commands based on the interaction objectives.

\begin{table*}[h]
    \centering
    \caption{\textbf{Quantitative comparison} of success rates, placement precision and execution velocity in Mujoco.}
    \label{tab:sim_comparison}
    \begin{tabular}{lcccccccccc}
        \toprule
        \multirow{2}{*}{\textbf{Method}} 
        & \multicolumn{4}{c}{\textbf{In Distribution}}
        & \multicolumn{4}{c}{\textbf{Out Distribution}} \\
        \cmidrule(lr){2-5} \cmidrule(lr){6-9}
        & R$_{\text{succ}}$(\%$\uparrow$) 
        & R$_{\text{prec}}$(cm$\downarrow$)
        & R$_{\text{task}}$(\%$\uparrow$)
        & V$_{\text{exec}}$(m/s$\uparrow$)
        & R$_{\text{succ}}$(\%$\uparrow$) 
        & R$_{\text{prec}}$(cm$\downarrow$)
        & R$_{\text{task}}$(\%$\uparrow$)
        & V$_{\text{exec}}$(m/s$\uparrow$) \\
        \midrule
        w/o SDF      & 97.90 & 7.54 & 55.76 & 0.29 & 82.52 & 8.44 & 64.26 & 0.436 \\
        w/o contact  & 94.99 & 7.86 & 31.13 & 0.29 & 88.43 & 8.28 & 71.65 & 0.41 \\
        PhysHSI~\cite{physhsi}
                     & 96.60 & 7.62 & \textbf{90.59} & \textbf{0.49} & 82.54 & 7.25 & 70.17 & \textbf{0.72} \\
        \rowcolor{gray!20}
        \textbf{Ours}
                     & 97.90 & 6.32 & 73.07 & 0.29
                     & 91.90 & \textbf{6.83} & 76.15
                     & 0.42 \\
        \rowcolor{gray!20}
        \textbf{Ours w/ FR.}
                     & \textbf{100.00} & \textbf{5.80} & 73.67 & 0.29
                     & \textbf{99.93} & 6.84 & \textbf{88.38} & 0.27 \\
        \bottomrule
    \end{tabular}
\end{table*}

\section{Evaluation}
In this section, we conduct a comprehensive evaluation of our proposed framework in both high-fidelity simulation and real-world environments. Our experiments aim to answer the following questions:
\begin{itemize}
    \item \textbf{Q1:} How does our root-guided policy perform on box carrying tasks?
    \item \textbf{Q2:} How does our policy perform on tracking the reference root trajectories and contact command?
    \item \textbf{Q3:} Does the system support complex, long-horizon, and dynamic tasks?
    \item \textbf{Q4:} Can the policy and perception system reliably transfer to physical hardware?
\end{itemize}

\subsection{How does our root-guided policy perform on box carrying tasks?}
\textbf{Experimental Setup:} The task is decomposed into four phases: \textit{Approach}, \textit{Pick}, \textit{Carry}, and \textit{Place}. To rigorously evaluate the robustness and generalization capability of our method, we conducted comprehensive tests categorized into two sets: \textit{In-Distribution (ID)} and \textit{Out-of-Distribution (OOD)}. The ID set utilizes the exact robot and object configurations from the training distribution. The OOD set is designed to test the policy by random sampling of the initial object position and the target placement position in the workspace. Specifically, we created a grid with a spacing of $0.2$m within a trapezoidal region defined by $x \in [0.4, 1.4]$ m and $y \in [-x, x]$ m. For each grid point, we evaluated object yaw orientations from $\{0^\circ, 15^\circ, 30^\circ, 45^\circ\}$. Furthermore, target placement locations were sampled along a circle with a radius of $4$m centered on the robot at $10^\circ$ intervals. This extensive evaluation protocol yields a total of $5,756$ distinct task scenarios.

Table \ref{tab:sim_comparison} presents the evaluation of our root-guide policy's performance by comparing it against the following baselines in the MuJoCo simulator:
\begin{itemize}
    \item \textbf{w/o SDF:} We trained a variant of our policy using raw retargeted motions without SDF optimization while keeping all other hyperparameters constant.
    \item \textbf{w/o contact:} This baseline is trained without the contact-related observation and reward.
    \item \textbf{PhysHSI:} A state-of-the-art baseline that utilizes AMP combined with manually designed rewards for box carrying tasks.
    \item \textbf{Ours w/ FR.:} A variant of our proposed framework that includes the failure recovery strategy.
\end{itemize}

The performance is evaluated using the following metrics:

\subsubsection{\textbf{Grasp Success Rate}}
\begin{figure*}[tp]
    \centering
    \begin{minipage}{0.65\linewidth}
        \centering
        \includegraphics[width=\linewidth]{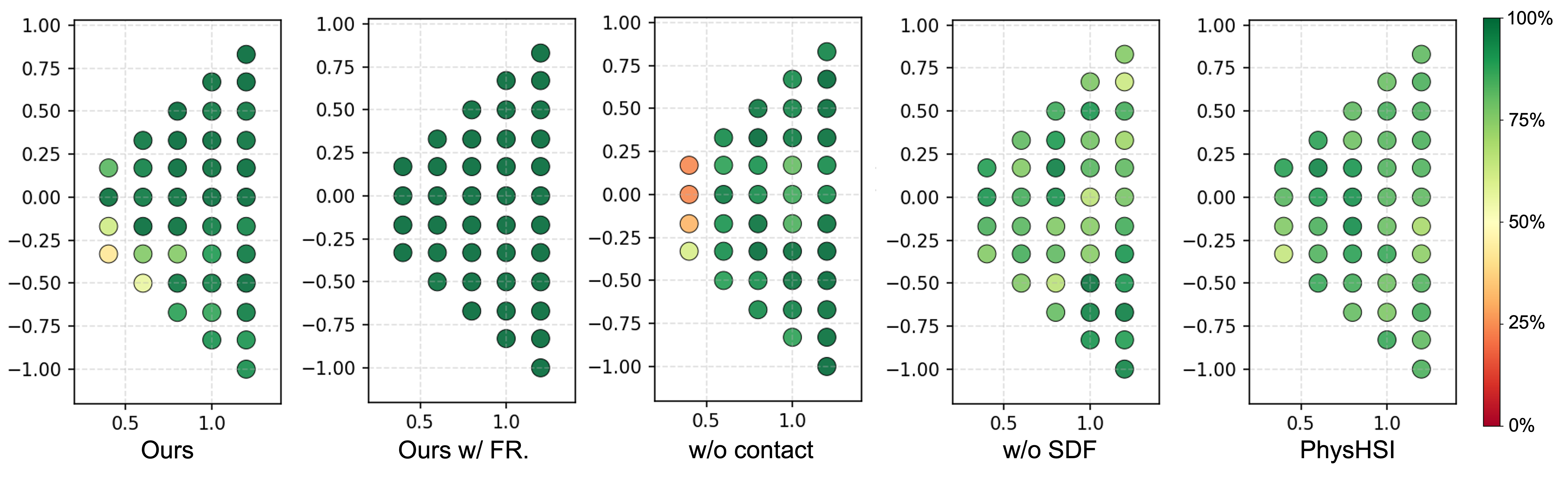}
        \captionof{figure}{\label{fig:grasp_heatmap}\textbf{Spatial distribution of grasp success rates across the Out-of-Distribution (OOD) evaluation set.} 
    The color scale indicates the success rate, where darker green denotes higher stability and red/yellow indicates failure. }
    \end{minipage}
    \hfill
    \begin{minipage}{0.32\linewidth}
        \centering
        \includegraphics[width=\linewidth]{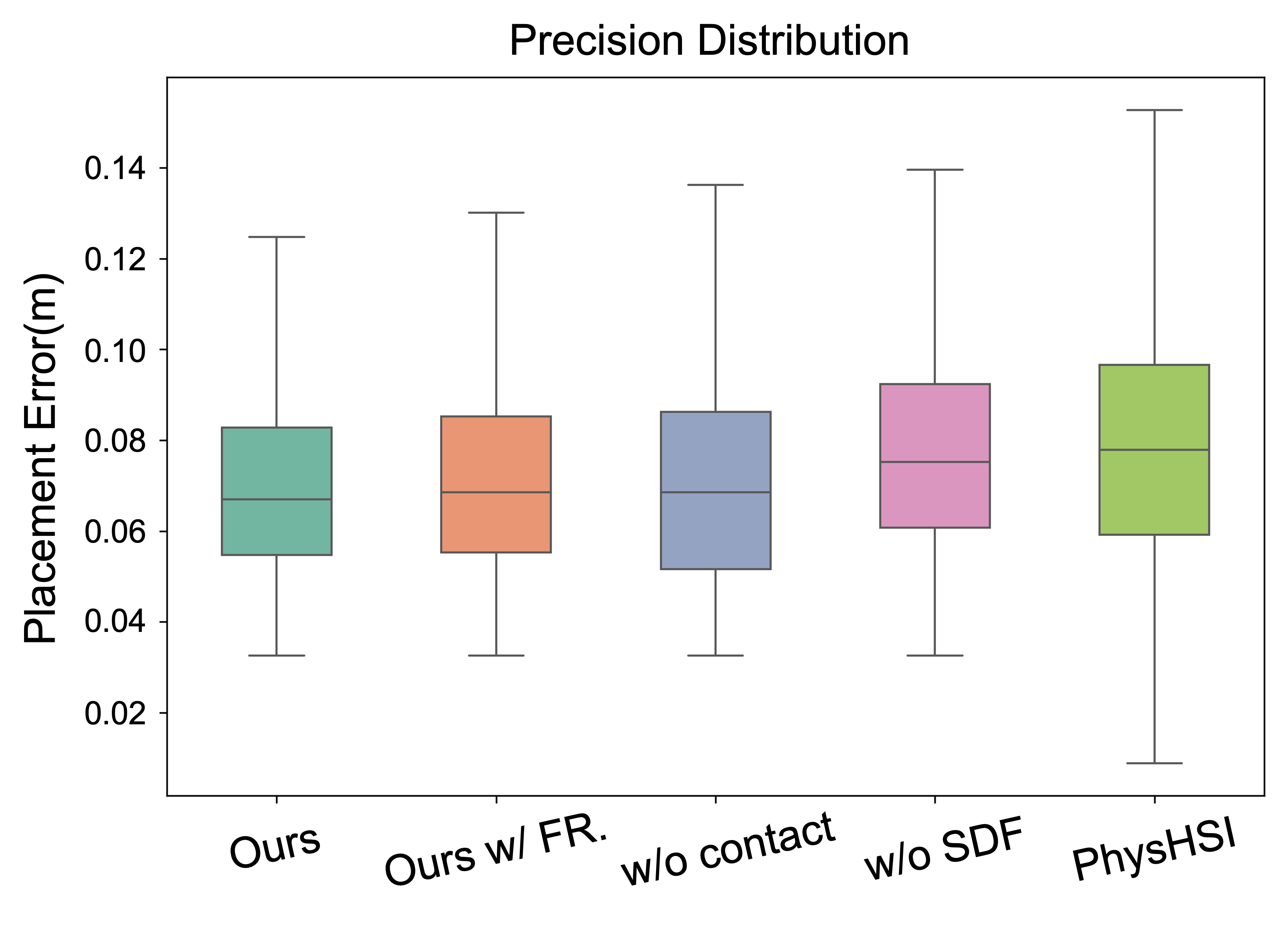}
        \captionof{figure}{\label{fig:place_precision}\textbf{Statistical distribution of placement precision across different methods.} }
    \end{minipage}
\vspace{-0.4cm}
\end{figure*}

Evaluates the stability of the grasp and represents it as ($R_\text{succ}$). It is defined as the percentage of trials where the robot successfully lifts the object above a threshold height and maintains the grasp without slippage for at least $1$ second.
Our method demonstrates superior grasping stability, particularly in generalization scenarios. While baseline methods suffer significant performance drops in OOD settings (e.g., PhysHSI drops from 96.60\% to 82.54\%), our full method (\textbf{Ours w/ FR.}) achieves a near-perfect success lift rate of 99.93\% in OOD tasks. This robust performance verifies that our root-guided policy, combined with the failure recovery mechanism, effectively handles variations in object positioning without compromising grasp stability. A detailed visualization of the grasp success rate is presented in Figure \ref{fig:grasp_heatmap}.

\subsubsection{\textbf{Placement Precision}}
Quantifies the manipulation accuracy and represents it as $R_{\text{prec}}$. It is calculated as the average Euclidean distance between the object's final resting position and the target goal center. This metric is computed only for successful episodes to exclude outliers from failed grasps.
Quantitatively, our approach yields the highest placement precision. In ID settings, \textbf{Ours w/ FR.} reduces the placement error to just 6.46 cm, significantly outperforming the baselines, which hover around 9.5 cm. This trend continues in OOD scenarios, where both our base policy and the version with failure recovery maintain high accuracy. This indicates that our minimalist observation space does not sacrifice control precision; rather, by focusing on the root trajectory, the policy achieves more consistent global spatial planning. We present the box plot of placement precision with tolerance threshold $< 0.2$m in Figure \ref{fig:place_precision}.

\subsubsection{\textbf{Task Success Rate}}
The metric for task completion rate is represented as $R_{\text{task}}$. A trial is considered a task success only if the robot completes the entire pick-and-place sequence, and the final object position lies within a tolerance threshold ($< 0.1$m) of the target.
The results highlight the generalization capability of our framework. Although PhysHSI achieves a high success rate (90.59\%) in ID settings, likely due to overfitting to reference motions, its performance degrades to 70.17\% in OOD tasks. In contrast, \textbf{Ours w/ FR.} demonstrates exceptional robustness, achieving the highest task success rate of 88.38\% in the challenging OOD setting. This validates that our decoupled observation design enables the robot to adapt to novel object locations and complete complex multi-stage tasks that traditional motion-tracking methods fail at.

\begin{table}[H]
    \centering
    \caption{Quantitative comparison of tracking fidelity and synchronization. Metrics are reported as \textbf{Mean $\pm$ Std}. \textbf{RPE}: Root Position Error (m), \textbf{ROE}: Root Orientation Error (rad), \textbf{MPL}: Motion Phase Lag (s), \textbf{CPL}: Contact Phase Lag (s). The best results are highlighted in \textbf{bold}.}
    \label{tab:tracking_metrics}
    \resizebox{\linewidth}{!}{
    \begin{tabular}{lcccc}
        \toprule
        \textbf{Method} & \textbf{RPE} $\downarrow$ & \textbf{ROE} $\downarrow$ & \textbf{MPL} $\downarrow$ & \textbf{CPL} $\downarrow$ \\
        \midrule
        w/o Contact & 0.23 \ci{0.03}  & 6.39 \ci{0.53}  & 18.37 \ci{1.54}  & 20.50 \ci{7.00}  \\
        w/o SDF & 0.26 \ci{0.03}  & 6.33 \ci{0.79}  & 22.11 \ci{1.35}  & \textbf{14.25 \ci{3.28} } \\
        \midrule
        \textbf{Ours} & \textbf{0.22 \ci{0.02} } & \textbf{5.77 \ci{0.41} } & \textbf{16.21 \ci{1.61} } & {14.80 \ci{3.89} } \\
        \bottomrule
    \end{tabular}
    }
\end{table}

\subsection{How does our policy perform on tracking the reference root trajectories and contact command?}

To rigorously quantify the policy's ability to track the reference root pose and adhere to the contact schedule, we conducted an extensive evaluation using diverse trajectory generation methods across a range of target velocities. We recorded both the generated reference trajectories, $\xi_{\text{ref}}$, and the robot's executed root trajectories, $\xi_{\text{real}}$. The performance is measured using the following four metrics:

\begin{itemize}
    \item \textbf{Root Position Error (RPE):} The average Euclidean distance between the reference and actual root positions.
    \item \textbf{Root Orientation Error (ROE):} The average angular deviation between the reference and actual root orientations.
    \item \textbf{Motion Phase Lag (MPL):} Defined as the temporal delay for the robot's root to reach the spatial position specified by the reference trajectory. This metric reflects the dynamic response speed of the policy.
    \item \textbf{Contact Phase Lag (CPL):} The average temporal discrepancy between the reference contact timestamp (when $c_t$ switches to 1) and the actual physical contact onset.
\end{itemize}

We benchmark our method against two ablation variants, \textbf{w/o SDF} and \textbf{w/o Contact} as described above.

The quantitative results presented in Table~\ref{tab:tracking_metrics} demonstrate that our full framework consistently outperforms the ablation baselines across most metrics, achieving the best balance between spatial accuracy and temporal synchronization.

In terms of spatial tracking fidelity, ours achieves the lowest RPE and ROE. Compared to the \textit{w/o SDF} baseline, our method significantly reduces tracking deviations, suggesting that the SDF-based optimization effectively eliminates physics inconsistencies in the reference motion, thereby allowing the policy to learn more precise kinematic tracking.

Regarding dynamic response, our method exhibits the lowest MPL, improving upon \textit{w/o Contact} and \textit{w/o SDF}. This indicates that including explicit contact states and high-quality motion priors enables the policy to anticipate high-dynamic transitions more effectively, thereby reducing the temporal delay between the reference and executed trajectories.

For contact synchronization CPL, the \textit{w/o Contact} variant exhibits the highest lag, confirming that without explicit contact observations, the policy struggles to initiate interaction at the correct time. While \textit{w/o SDF} achieves a marginally lower CPL  compared to \textbf{Ours}, the difference is minimal given the standard deviation. Crucially, our method maintains this competitive contact timing while delivering significantly superior spatial tracking and motion response, validating the necessity of each component in the proposed framework.

\subsection{Does the system support complex, long-horizon, and dynamic tasks?}
To demonstrate the versatility and robustness of our integrated system, we designed four challenging task scenarios as shown in Figure \ref{fig:teaser}:
\paragraph{Task I: Continuous Long-Horizon Transport}
We verified the system's ability to perform continuous re-arrangement tasks. Guided by the root trajectory planner, the robot autonomously returns to the starting point after placing the box, initiating the next transport cycle. This test confirms the cyclic stability of our state machine and planner.
\paragraph{Task II: Obstacle Avoidance}
We introduced static obstacles in the transport path. Leveraging the TEB planner, our system successfully generated collision-free trajectories, modifying the robot's yaw and path to navigate around obstacles while carrying the load.
\paragraph{Task III: Failure Recovery via Digital Twin}
We simulated a scenario where the box is knocked out of the robot's hands during transport. The Object State Estimation module detected the drop and utilized the physics-based simulator to estimate the object's new location. The robot successfully adjusted its gaze, replanned the approach, and regrasped the object, demonstrating resilience in the face of unexpected failures.
\paragraph{Task IV: High-Dynamic Agility}
To explore the upper limits of our method, we simulated a high-speed "running and carrying" scenario. The results indicate that our root-guided policy can generalize to highly dynamic motions, suggesting potential for future high-speed humanoid manipulation.
\subsection{Can the policy and perception system reliably transfer to physical hardware?}
We deployed our system on the Unitree G1 humanoid robot to evaluate sim-to-real transferability.
\subsubsection{Real-World Manipulation and Dynamic Robustness}
We deployed our system on the physical Unitree G1 robot to evaluate its robustness under varying dynamic conditions. To systematically assess the policy's performance across different operating speeds, we categorized the guidance commands into three profiles: \textit{Slow} ($v=0.2$ m/s, $\omega=0.6$ rad/s), \textit{Middle} ($v=0.4$ m/s, $\omega=1.0$ rad/s), and \textit{Fast} ($v=0.6$ m/s, $\omega=1.3$ rad/s).

The quantitative results are summarized in Table~\ref{tab:real_world_dynamic}. As shown, the system exhibits consistent reliability across diverse velocity profiles, maintaining a grasp success rate exceeding 60\% in all scenarios. We specifically conducted an extended evaluation of 28 consecutive trials under the \textit{Middle} profile. Notably, this sequence included multiple instances where the object was intentionally dislodged. Despite these severe disturbances, our method successfully recovered and completed the tasks, underscoring the system's exceptional robustness and stability on physical hardware.

\begin{table}[H]
    \centering
    \caption{Real-world performance under different dynamic command profiles. \textbf{GSR}: Grasp Success Rate (\%), \textbf{PP}: Placement Precision (error in m). The velocities represent the maximum commanded linear ($v$) and angular ($\omega$) limits.}
    \label{tab:real_world_dynamic}
    \begin{tabular}{lcccc}
        \toprule
        \multirow{2}{*}{\textbf{Profile}} & \multicolumn{2}{c}{\textbf{Command Velocity}} & \multirow{2}{*}{\textbf{GSR} ($\uparrow$)} & \multirow{2}{*}{\textbf{PP} (m) ($\downarrow$)} \\
        \cmidrule(lr){2-3}
         & $v$ (m/s) & $\omega$ (rad/s) & & \\
        \midrule
        \textbf{Slow} & 0.2 & 0.6 & 7 / 10 & 0.15\\
        \textbf{Middle} & 0.4 & 1.0 & 21 / 28 & 0.16 \\
        \textbf{Fast} & 0.6 & 1.3 & 6 / 10 & 0.25\\
        \bottomrule
    \end{tabular}
\end{table}

\subsubsection{Evaluation of Object State Estimation}
\begin{figure}[h]
    \centering
    \includegraphics[width=\linewidth]{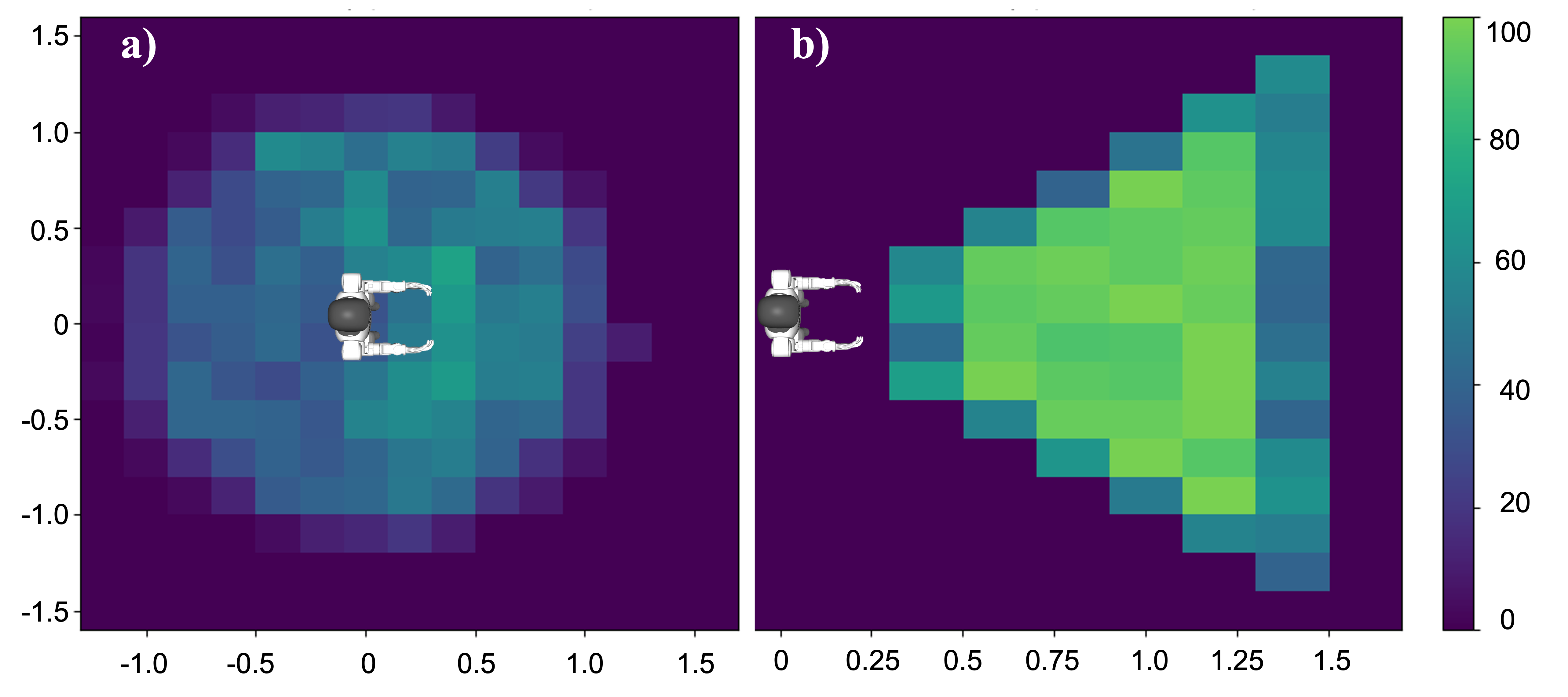}
    \caption{The percentages quantify the likelihood of successfully detecting an object dropped within each respective region. (a) The detection probability distribution simulated via the object estimate module. (b) The baseline detection probability distribution relying exclusively on the onboard camera.}
    \label{fig:drop_heatmap}
\end{figure}

To validate the efficacy of our Digital Twin drop prediction module, we established a discretized grid coordinate system centered on the robot's base with a spatial resolution of $0.2 \times 0.2$ m. Controlled drop experiments were conducted by releasing objects with arbitrary orientations within the robot's FoV. The primary function of this module is to estimate a recovery pose by falling simulation in MuJoCo and reorient the robot's gaze after the object fell. Consequently, we define the success rate for each grid cell as the Success rate with which the robot re-acquires the object in its FoV after adjusting its pose based on the drop location predicted by the MuJoCo simulation. Figure~\ref{fig:drop_heatmap} visualizes the detection success rates within the original camera FoV and the adjusted FoV guided by the object estimation. The results demonstrate that this module effectively augments the original detection FoV during the interaction process, especially the circular area centered on the robot.

\section{Conclusion} 
\label{sec:conclusion}

In this work, we presented Pro-HOI (Perceptive Root-guided Humanoid-Object Interaction), a generalizable and robust framework for humanoid-object interaction. By proposing a low-dimensional representation that encodes the interaction process through root trajectories and contact information, our approach overcomes the inherent generalization limitations of traditional imitation based learning method while make the HOI process controllable. Extensive experiments demonstrate that Pro-HOI achieves superior task success rates in out-of-distribution scenarios, enhanced manipulation precision, and reliable sim-to-real transferability compared to state-of-the-art baselines.

\section*{Acknowledgments}
This work is supported by the National Natural Science Foundation of China (Grant No.62306242),  the Young Elite Scientists Sponsorship Program by CAST (Grant No. 2024QNRC001), and the Yangfan Project of the Shanghai (Grant No.23YF11462200).

\bibliographystyle{plainnat}
\bibliography{references}

\clearpage

\begin{center}
\Large\bfseries Supplementary Appendix
\end{center}

\begin{figure*}[b]
    \centering
    \includegraphics[width=0.95\linewidth]{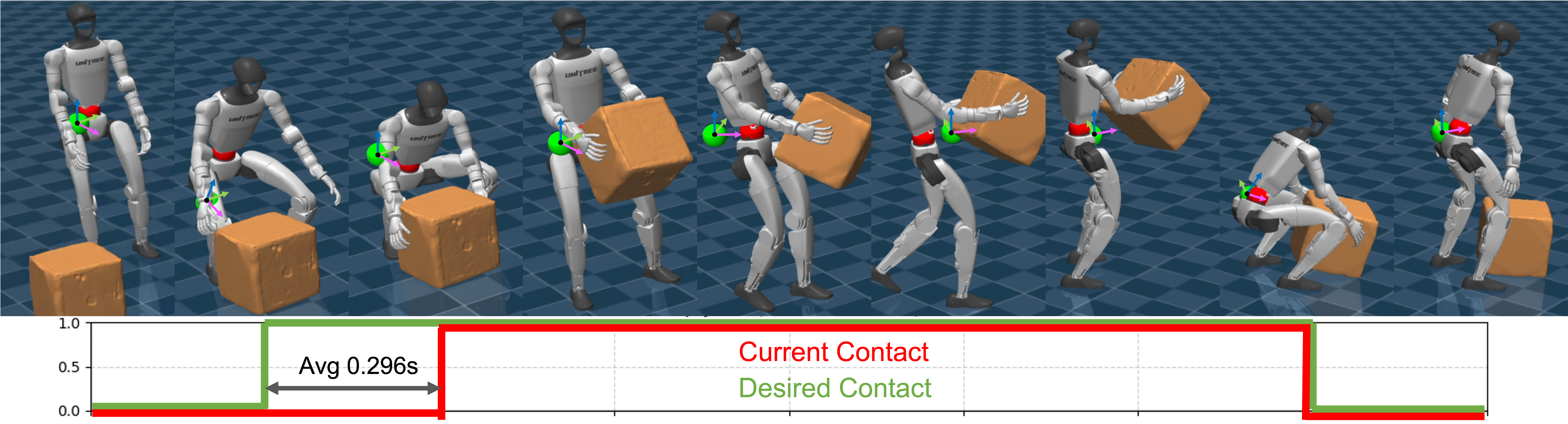}
    \caption{\label{fig:contact_phase}The example of trajectory generation with contact information and simulated tracking contact.}
\end{figure*}

\section{Detail of Implementation}
\subsection{Training and Simulation Setup} 
We utilized NVIDIA Isaac Lab as the reinforcement learning simulation environment. The policy training was conducted on a high-performance computing platform equipped with eight NVIDIA RTX 3090 GPUs and dual AMD CPUs. For simulation-based validation experiments in MuJoCo, we utilized an Apple MacBook Pro powered by the M3 silicon chip.

\subsection{Perception Detail}
The FastLIO2 robot state estimator, FoundationPose object pose estimator, and the robot controller are fully deployed on the robot's onboard NVIDIA Jetson Orin NX module. For real-time perception, we used the isaac-ros framework to interface the foundationpose model and trained the YOLOv8 model for the object. The YOLO results are converted to a mask, which is supplied to FoundationPose at the initial time. The object model is constructed by iPhone14 Pro using AR code Object Capture and converted to OBJ format with online tools. An external desktop workstation receives real-time estimates of the robot's state for visualization and logging. Communication between the high-level motion controller and the Unitree low-level control interface is facilitated via the Lightweight Communications and Marshalling (LCM) protocol. Ultimately, we obtained object perception at approximately 40 Hz, robot state estimation at 200 Hz, and robot control at 50 Hz.

\subsection{Detail of Object State Estimation Module}
The drop detection logic relies on a spatial safety threshold $\delta_{\text{pos}}$ and a temporal window $N$. Specifically, we define the safety region as a rectangular prism in the robot's base frame, with dimensions of $0.4 \times 0.2 \times 0.4$ m (length $\times$ width $\times$ height) and centered at $x=0.35$ m along the robot's forward axis. The sliding window size for consistent state monitoring is set to $N=10$. The Digital Twin simulation is hosted on an external desktop workstation within the same local area network. It receives the robot's object state stream via the ZeroMQ protocol. The module computes the object's velocity in real time; upon triggering a drop event, the initial velocity for the physics simulation is derived by averaging the velocities from the last 3 frames to mitigate noise. Once the simulator predicts the final resting position, the module calculates the required gaze-adjustment pose and transmits it to the robot for recovery.

\subsection{Detail of Root Trajectory Planner}
In this appendix, we provide the detailed mathematical formulation for the hybrid root trajectory planner described in Section \ref{subsubsec:root_planner}. This includes the quantitative retrieval metric for interaction priors, the explicit transformation from 2D navigation paths to 6-DoF root references, and the interpolation sequence construction.

\subsubsection{Interaction Pose Prior}To retrieve the optimal grasping prior, we first formalize the spatial poses using rigid transformation matrices $\mathbf{T} \in SE(3)$, where each $\mathbf{T}$ encapsulates a rotation $\mathbf{R} \in SO(3)$ and a translation $\mathbf{p} \in \mathbb{R}^3$. Let $\mathcal{P} = \{ (\mathbf{T}_{\text{obj}}^{(i)}, \mathbf{T}_{\text{int}}^{(i)}) \}_{i=1}^N$ denote the library of prior poses extracted from the training dataset. These represent the object pose $\mathbf{T}_{\text{obj}}^{(i)}$ and the corresponding robot interaction pose $\mathbf{T}_{\text{int}}^{(i)}$ (relative to the initial frame's robot coordinate system) recorded at the onset of the contact phase. Given a newly detected relative object pose $\mathbf{T}_{\text{obj}}^{\text{cur}}$, we define a weighted error function $S_i$ to quantify the similarity between the current state and the stored priors:
\begin{equation}
    S_i = w_T \| \mathbf{p}_{\text{obj}}^{\text{cur}} - \mathbf{p}_{\text{obj}}^{(i)} \|_2 + w_R \cdot \Phi(\mathbf{R}_{\text{obj}}^{\text{cur}}, \mathbf{R}_{\text{obj}}^{(i)})
\end{equation}
where $\Phi(\cdot)$ calculates the angular distance, and $w_T, w_R$ are weighting coefficients for translation and rotation errors, respectively. The system identifies the optimal prior index via $k = \operatorname*{argmin}_i S_i$. Subsequently, the target interaction pose is computed by compensating for the deviation between the current and prior object locations:
\begin{equation}
    \mathbf{p}_{\text{target}} = \mathbf{R}_{\text{root}}^z \left( \mathbf{p}_{\text{int}}^{(k)} + (\mathbf{p}_{\text{obj}}^{\text{cur}} - \mathbf{p}_{\text{obj}}^{(k)}) \right) + \mathbf{p}_{\text{root}}
\end{equation}
\begin{equation}
    \mathbf{R}_{\text{target}} = \mathbf{R}_{\text{root}} \mathbf{R}_{\text{int}}^{(k)}
\end{equation}
This target is then transformed into the world frame based on the robot's current root state to guide the policy.

\subsubsection{2D-to-3D Waypoint Lifting}For long-horizon navigation, the TEB local planner generates a planar trajectory $\mathcal{T}_\text{2D} = \{ (x_i, y_i, \psi_i) \}_{i=1}^M$, where $(x_i, y_i)$ denotes the 2D position and $\psi_i$ the yaw angle. To adapt this for our whole-body controller, we lift these 2D waypoints into a 6-DoF reference trajectory $\mathcal{T}_\text{waypoints} = \{ \mathbf{T}_\text{ref}^{(i)} \}_{i=1}^M$ using the following formulation:
\begin{equation}
    \mathbf{p}_\text{ref}^{(i)} = [x_i, y_i, h_\text{stand}]^\top
\end{equation}
\begin{equation}
    \mathbf{R}_\text{ref}^{(i)} = \mathbf{R}_z(\psi_i)
\end{equation}
Here, $h_\text{stand}$ represents the robot's default standing height, and $\mathbf{R}_z(\psi_i)$ is the rotation matrix corresponding to the yaw angle $\psi_i$, assuming zero roll and pitch to maintain stable locomotion.

\subsubsection{Trajectory Interpolation Sequence}To ensure smooth transitions between navigation and manipulation phases, we construct a continuous time-parameterized trajectory $\xi(t)$ using 6-DoF interpolation. The interpolation sequence explicitly includes intermediate "via-points" adjusted to the standing height ($\mathbf{T}^{\text{up}}$) to prevent collision during approach and transport:
\begin{equation}
\begin{aligned}
\xi(t) = \operatorname{Interp}\bigl(
    &\mathbf{T}^{\text{curr}},
    \mathbf{T}^{\text{up}}_{\text{target}}, \mathbf{T}_{\text{target}},
    \mathbf{T}^{\text{up}}_{\text{target}}, \mathbf{T}_{\text{ref}}^{(1)}, \dots, \\
    &\mathbf{T}_{\text{ref}}^{(M)},
    \mathbf{T}^{\text{up}}_{\text{end}}, \mathbf{T}_{\text{end}},
    \mathbf{T}^{\text{up}}_{\text{end}}
\bigr)
\end{aligned}
\end{equation}
The contact information $c_t$ is manually set to 1 during the interval $[\mathbf{T}_{\text{target}}, \mathbf{T}_{\text{end}}]$, and 0 otherwise. The example is shown in Figure \ref{fig:contact_phase}. This sequence ensures the robot lifts its root to a safe height before and after interacting with the target $\mathbf{T}_{\text{target}}$ or the final goal $\mathbf{T}_{\text{end}}$.

\section{Supplementary Tables}
\label{sec:rew}
This section presents four tables detailing the implementation of our Pro-HOI framework. These tables specify the weights of the reward terms, the dimensions of the observation space, the domain randomization settings, and the hyperparameters used for training.

\begin{table*}[ht]
    \centering
    \label{tab:ppo_reward}

    \renewcommand{\arraystretch}{1.2}
    {\fontsize{8}{8}\selectfont

    \begin{minipage}[t]{0.3\textwidth}
        \centering
        \captionof{table}{Hyperparameters related to PPO.}
        \label{tab:ppo}

        \begin{tabular}{lc}
            \toprule
            Hyperparameter & Value \\
            \midrule
            Optimizer & Adam \\
            Batch size & 4096 \\
            Mini Batches & 4 \\
            Learning epoches & 5 \\
            Entropy coefficient & 0.01 \\
            Value loss coefficient & 1.0 \\
            Clip param & 0.2 \\
            Max grad norm & 1.0 \\
            Init noise std & 1.0 \\
            Learning rate & 1e-3 \\
            Desired KL & 0.01 \\ 
            GAE $\lambda$ & 0.95\\
            GAE $\gamma$ & 0.99\\
            Actor MLP & [512,256,128] \\
            Critic MLP & [512,256,128] \\
            Activation & ELU \\
            \bottomrule
        \end{tabular}
    \end{minipage}
    \hfill
    \begin{minipage}[t]{0.68\textwidth}
        \centering
        \captionof{table}{Reward terms and weights.}
        \label{tab:reward_desgin}

        \renewcommand{\arraystretch}{1.35}

        \begin{tabular}{lcc}
            \toprule
            Term & Expression & Weight \\
            \midrule
            \rowcolor[HTML]{EFEFEF} & \multicolumn{1}{c}{Task} &\\ 
            \midrule

            Global robot position &
            $\exp( -\|\bm{p}_t-\hat{\bm{p}}_t\|^2/\sigma_{\mathrm{rpos}} )$
            & 0.5 \\

            Global robot orientation &
            $\exp(- \|\bm{r}_t-\hat{\bm{r}}_t\|^2/\sigma_{\mathrm{rrot}})$
            & 0.5 \\

            Global body position &
            $\exp(-\|\bm{p}^{b}_t-\hat{\bm{p}}^{b}_t\|^2/\sigma_{\mathrm{bpos}})$
            & 1.0 \\

            Global body orientation &
            $\exp(-\|\bm{r}^{b}_t-\hat{\bm{r}}^{b}_t\|^2/\sigma_{\mathrm{brot}})$
            & 1.0 \\

            Body linear velocity &
            $\exp(-\|\bm{v}^{b}_t-\hat{\bm{v}}^{b}_t\|^2/\sigma_{\mathrm{bvel}})$
            & 1.0 \\

            Body angular velocity &
            $\exp(-\|\bm{\omega}^{b}_t-\hat{\bm{\omega}}^{b}_t\|^2/\sigma_{\mathrm{bang}})$
            & 1.0 \\

            Object position &
            $\exp(-\|\bm{p}^{o}_t-\hat{\bm{p}}^{o}_t\|^2/\sigma_{\mathrm{opos}})$
            & 1.0 \\

            \midrule
            \rowcolor[HTML]{EFEFEF} & \multicolumn{1}{c}{Regularization} &\\ 
            \midrule

            Joint limits &
            $\mathbb{I}(\bm{q}\notin[\bm{q}_{\min},\bm{q}_{\max}])$
            & -10.0 \\

            Undesired contacts &
            $\sum \mathbb{I}(\|\mathbf{F}\|>1.0)$
            & -0.1 \\

            Action rate &
            $\|\bm{a}_t-\bm{a}_{t-1}\|^2$
            & -0.3 \\

            \bottomrule
        \end{tabular}
    \end{minipage}

    }

\end{table*}

\subsection{Reward Terms and Weights}
All reward functions are detailed in Table~\ref{tab:reward_desgin}. The overall reward design is divided into two categories: tracking rewards and regularization rewards. The \textit{undesired contacts} term is employed to penalize self-collisions between the robot’s links. Specifically, the set $\mathcal{B}$ includes all robot bodies, with the exception of the ankles and wrists. The relative body includes the pelvis, hip, knee, ankle, shoulder, elbow, wrist, and torso links.

\subsection{Observation}
\label{sec:obs}
The specific observation items and dimensions for the Actor and Critic are detailed in Table~\ref{tab:obs}.
\begin{table}[h]
    \centering
    \caption{Actor and critic observation state space.}
    \label{tab:obs}
    \renewcommand{\arraystretch}{1.1}
    {\fontsize{8}{8}\selectfont
    \begin{tabular}{lcc}
        \toprule
        State term & Actor Dim & Critic Dim \\
        \midrule
        Global root position & 3 & 3 \\
        Global root orientation & 6 & 6 \\
        Relative object position & 3 & 3 \\
        Relative object orientation & 6 & 6 \\
        Contact information & 1 & 1\\
        Root projected gravity & 3  & 3 \\
        Root angular velocity & 3 & 3 \\
        Joint position & 29 & 29 \\
        Joint velocity & 29 & 29 \\
        Actions & 29 & 29 \\
        \midrule
        Root linear velocity & -- & 3 \\
        Relative Body position & -- & 45 \\
        Relative Body orientation & -- & 90 \\
        \midrule
        \textbf{Total dim} & 112 & 250 \\
        \bottomrule
    \end{tabular}
    }
    
\end{table}

\subsection{Domain Randomization} 
Since the physical characteristics of the simulator cannot accurately represent the real situation, we incorporate domain randomization during training to improve the transferability of trained polices to real-world or other simulations settings. The specific settings are given in Table \ref{tab:domain_random}. 
\label{sec:dr}

\subsection{PPO Hyperparameter}
The detailed PPO hyperparameters are shown in Table \ref{tab:ppo}. 
\label{sec:ppo}

\begin{table*}[!t]
\vspace{-180mm}
\centering
\footnotesize
\setlength{\tabcolsep}{6pt}
\begin{tabular}{l l}
\toprule
\textbf{Domain Randomization} & \textbf{Sampling Distribution} \\
\midrule
\multicolumn{2}{l}{\textit{Physical parameters}} \\
\quad Static friction coefficients & $\mu_s \sim \mathcal{U}[0.3,1.6]$ \\
\quad Dynamic friction coefficients & $\mu_d \sim \mathcal{U}[0.3,1.2]$ \\
\quad Restitution coefficient & $e \sim \mathcal{U}[0,0.5]$ \\
\quad Default joint positions & $\bm{q}_0 \sim \bm{q}_0 + \mathcal{U}[-0.01, 0.01]$ \\
\quad Base COM offset (x, y, z) & $\Delta x \sim \mathcal{U}[-0.025,0.025],\ \Delta y \sim \mathcal{U}[-0.05,0.05],\ \Delta z \sim \mathcal{U}[-0.05,0.05]$ \\

\multicolumn{2}{l}{\textit{Root velocity perturbations (external pushes)}} \\
\quad Root linear vel (x, y, z) & $v_x \sim \mathcal{U}[-0.5,0.5],\ v_y \sim \mathcal{U}[-0.5,0.5],\ v_z \sim \mathcal{U}[-0.2,0.2]$ \\
\quad Push duration & $\Delta t \sim \mathcal{U}[1,3]\text{s}$ \\
\quad Root angular vel & $\omega_{\text{roll}} \sim \mathcal{U}[-0.52,0.52],\ \omega_{\text{pitch}} \sim \mathcal{U}[-0.52,0.52],\ \omega_{\text{yaw}} \sim \mathcal{U}[-0.78,0.78]$ \\

\multicolumn{2}{l}{\textit{Task related}} \\
\quad Object mass & $\text{scale}_o \sim \mathcal{U}[0.5,2.0]$ \\

\bottomrule
\end{tabular}
\caption{Domain randomization parameters applied during training. $\mathcal{U}[\cdot]$: uniform distribution.}
\label{tab:domain_random}
\end{table*}

\end{document}